\pdfoutput=1

\documentclass[11pt]{article}

\usepackage{naacl2021}

\usepackage{times}
\usepackage{latexsym}
\usepackage{xspace}

\usepackage[T1]{fontenc}

\usepackage[utf8]{inputenc}

\usepackage{microtype}

\usepackage{comment}
\usepackage{amsmath,amssymb,amsthm}
\usepackage{url}
\usepackage{multirow}
\usepackage{graphicx}
\usepackage{subcaption,siunitx,booktabs}
\usepackage{tabularx,comment}
\usepackage{makecell}
\usepackage{arydshln}
\usepackage{multirow}
\usepackage{float}

\newcommand{\ours}{DART\xspace}

\title{\ours: Open-Domain Structured Data Record to Text Generation}

\author{
Linyong Nan$^{1}$
\quad Dragomir Radev$^{1,2}$
\quad Rui Zhang$^{3}$
\quad Amrit Rau$^{1}$
\quad Abhinand Sivaprasad$^{1}$
\\{\bf
\quad Chiachun Hsieh$^{4}$
\quad Xiangru Tang$^{1}$
\quad Aadit Vyas$^1$
\quad Neha Verma$^1$
\quad Pranav Krishna$^5$
}
\\{\bf
\quad Yangxiaokang Liu$^1$
\quad Nadia Irwanto$^1$
\quad Jessica Pan$^1$
\quad Faiaz Rahman$^1$
\quad Ahmad Zaidi$^1$
}
\\{\bf
\quad Murori Mutuma$^1$
\quad Yasin Tarabar$^1$
\quad Ankit Gupta$^1$
\quad Tao Yu$^1$
\quad Yi Chern Tan$^1$
\quad 
}
\\{\bf
\quad Xi Victoria Lin$^2$\thanks{\ \ Now at Facebook AI.}
\quad Caiming Xiong$^2$
\quad Richard Socher$^2$
\quad Nazneen Fatema Rajani$^2$
}
\\
$^1$ Yale University \quad $^2$ Salesforce Research \quad $^3$ Penn State University\\ 
$^4$ The University of Hong Kong \quad $^5$ MIT\\
\small{\tt{\{linyong.nan, dragomir.radev\}@yale.edu}, \tt{rmz5227@psu.edu}, \tt{nazneen.rajani@salesforce.com}}
}

\begin{document}
\maketitle
\begin{abstract}
We present \textbf{\ours}, an open domain structured \textbf{DA}ta \textbf{R}ecord to \textbf{T}ext generation dataset with over 82k instances (DARTs).
Data-to-Text annotations can be a costly process, especially when dealing with tables which are the major source of structured data and contain non-trivial structures. To this end, we propose a procedure of extracting semantic triples from tables that encodes their structures by exploiting the semantic dependencies among table headers and the table title.
Our dataset construction framework effectively merged heterogeneous sources from open domain semantic parsing and dialogue-act-based meaning representation tasks by utilizing techniques such as: tree ontology annotation, question-answer pair to declarative sentence conversion, and predicate unification, all with minimum post-editing.
We present systematic evaluation on \ours~ as well as new state-of-the-art results on WebNLG 2017 to show that \ours~(1) poses new challenges to existing data-to-text datasets and (2) facilitates out-of-domain generalization. Our data and code can be found at \url{https://github.com/Yale-LILY/dart}.
\end{abstract}

\section{Introduction}
Automatically generating textual descriptions from structured data improves the accessibility of knowledge bases to lay users. Such applications include explaining data records to non-experts~\cite{cawsey1997natural}, writing sports news \cite{chen2008learning}, summarizing information in multiple documents \cite{fan2019using}, and generating dialogue responses \cite{wen2015semantically}.

While significant progress has been made in this field, there are still several issues with existing Data-to-Text datasets.
First, they adopt a flat ontology structure of the data, such as slot-value pairs for data records \cite{lebret2016neural,novikova2017e2e} or flat schema for tables~\cite{wiseman2017challenges,chen2020logical,parikh2020totto}.
This flat structure is not powerful enough to encode rich semantic relationships in the ontology of the structured data, especially tables, whose representation can be further improved with these semantic knowledge.
Second, some of the datasets only focus on a small number of domains or knowledge graphs, therefore providing limited number of predicates and data ontologies. For example, E2E \cite{novikova2017e2e} on restaurants and WebNLG \cite{gardent2017webnlg} on 15 categories from DBPedia.
Furthermore, some of them only have loose alignments between data input and sentence due to the nature of the task \cite{wiseman2017challenges} and the automatic generation procedure \cite{vougiouklis2018wikipedian, elsahar2018rex}.

\begin{figure*}[t]
  \centering
  \includegraphics[width=0.9\textwidth]{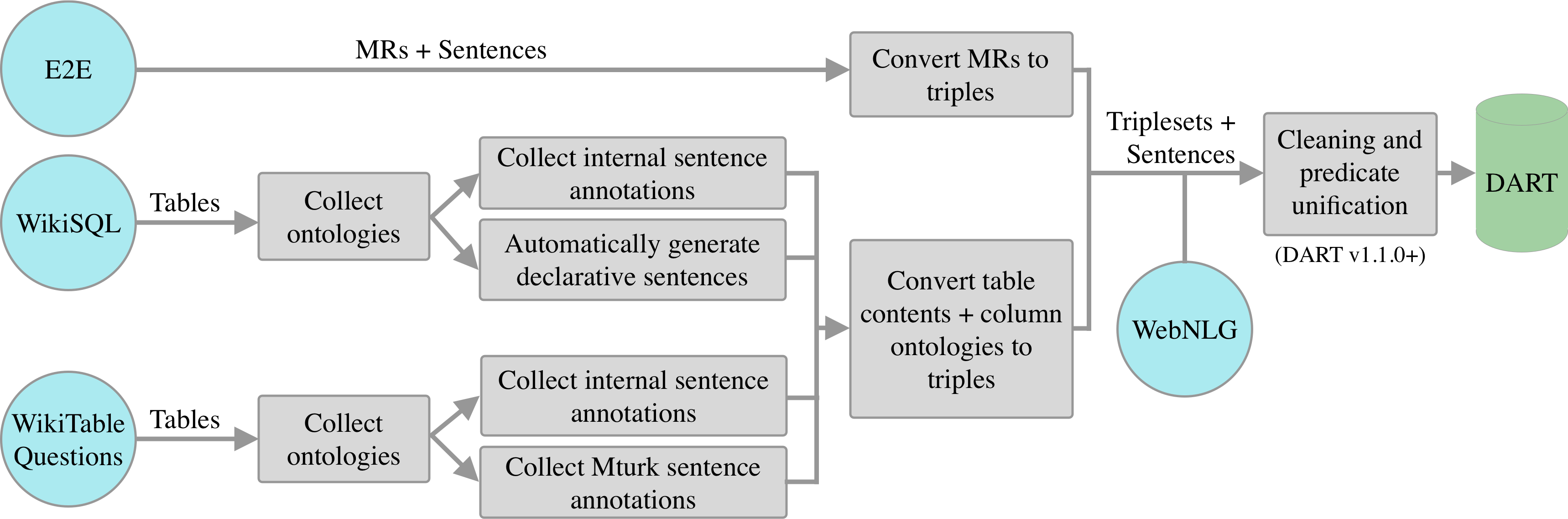}
  \caption{\ours~data collection pipeline. MR: Meaning Representation.}
  \label{fig:data_pipeline}
\end{figure*}

To address some of these issues and to encourage further research in natural language generation from structured data, we introduce \textbf{\ours}, a large and open-domain structured \textbf{DA}ta-\textbf{R}ecord-to-\textbf{T}ext generation corpus. The goal of DART is to harvest the diverse predicates occurred in Wikipedia tables, which is significantly richer than those defined in the domain specific ontologies E2E and WebNLG were built on (Table \ref{tab:dart}).
We also introduce novel tree ontology annotation on tables, which converts a flat table schema into a tree structured semantic frame.
The tree ontology reflects the core and auxiliary relations in the table schema, and naturally occurs across many domains. 
As a result, DART provides high-quality sentence annotations to tree structured semantic frames extracted from various data sources, including WikiSQL \cite{zhong2017seq2sql} and WikiTableQuestions \cite{pasupat2015compositional}, two open-domain question answering datasets, as well as E2E \cite{novikova2017e2e} and WebNLG \cite{gardent2017webnlg} (Figure \ref{fig:data_pipeline}).
We evaluated several state-of-the-art data-to-text models on \ours, and found that while these models achieve impressive performance on domain-specific datasets, their performance suffers on \ours~due to its open-domain nature and richer semantic structures.

Our contributions are as follows. (1) We present a large and open-domain corpus for structured data record to text generation, annotated with tree ontologies converted from the table. This \textit{hierarchical} input differentiates our corpus from existing data-to-text corpora. (2) We benchmark several state-of-the-art data-to-text models to show that DART introduces new generalization challenges. (3) We demonstrate that using DART for data augmentation improves the performance of existing models on the WebNLG 2017 dataset. We expect the results to generalize to other data-to-text datasets given the open-domain nature of DART.

\begin{table*}[ht!]
\centering
\resizebox{.85\textwidth}{!}{%
\begin{tabular}{@{}lccccccc@{}}
\toprule
                                  & Input Unit & Examples & Vocab Size  & Words per SR & Sents per SR & Tables \\
\midrule
WikiTableText  & Row               & 13,318    & ---         & 13.9         & 1.0    &  4,962      \\
LogicNLG       & Table             & 37,015    & 122K        & 13.8         & 1.0    &  7,392      \\
ToTTo          & Highlighted Cells & 136,161   & 136K        & 17.4         & 1.0    & 83,141     \\
\ours          & Triple Set        & 82,191    & 33.2K       & 21.6         & 1.5    &  5,623      \\
\bottomrule
\end{tabular}%
}
\caption{\ours~compared with other open-domain table-to-text datasets. \ours~takes triple sets as input by incorporating the ontology of table headers and title, and its surface realizations tend to be longer with more than single sentence verbalization. SR: Surface Realization.}
\label{tab:dart_compare}
\end{table*}

\section{\ours~Data Collection}

As shown in Figure \ref{fig:data_pipeline}, DART is constructed from three different sources: (1) human annotation on Wikipedia tables from two table semantic parsing and question answering datasets WikiSQL and WikiTableQuestions (\S~\ref{sec:human_annotation}), (2) automatic conversion of questions in WikiSQL to declarative sentences (\S~\ref{sec:declarative}), and (3) incorporation of existing datasets including WebNLG 2017 and Cleaned E2E (\S~\ref{sec:existing}).
After collecting the $\langle$triple-set, sentence$\rangle$ pairs from various data sources, we manually canonicalized the predicates and show that DART covers a broad range of topics (\S~\ref{sec:unification}).
Finally, we discuss the data split in \S~\ref{sec:dataset_split}.

\begin{table*}[ht!]
\centering
\resizebox{.88\textwidth}{!}{%
\begin{tabular}{lcccccc}
\toprule
                                       & \multicolumn{6}{c}{\ours: 62,659 train / 6,980 dev / 12,552 test} \\ \midrule
                                       & \multicolumn{2}{c}{WikiTableQuestions} & \multicolumn{2}{c}{WikiSQL} & \multicolumn{1}{c}{\multirow{2}{*}{WebNLG}} & \multicolumn{1}{c}{\multirow{2}{*}{Cleaned E2E}} \\
                                       & Internal           & MTurk             &  Internal   & Declarative      & \multicolumn{1}{c}{}                        & \multicolumn{1}{c}{}                             \\ \midrule
Domains                                & \multicolumn{4}{c}{Wikipedia (open-domain)}                          & 15 DBPedia Categories                       & Restaurants                                      \\
Unique Predicates                      & 1,950              & 1,403             & 493            & 2,008      & 347                                         & 7                                                \\
Unique Triples                         & 13,505             & 5,541             & 1,648          & 7,787      & 3,220                                       & 946                                              \\
Tripleset-Sentence Pairs               & 4,902              & 2,120             & 772            & 4,204      & 27,731                                      & 42,462                                           \\
Triples per Tripleset (min, med, max)  & 1, 3, 10           & 1, 3, 7           & 1, 2, 7        & 1, 2, 10   & 1, 3, 7                                     & 1, 4, 7                                          \\
Vocab Size                             & 13.4K              & 8.9K              & 3.0K           & 10.7K      & 8.0K                                        & 3.0K                                             \\
Words per SR                      & 15.2               & 16.5              & 14.0           & 12.6       & 22.5                                        & 22.9                                             \\
Sentences per SR                  & 1.0                & 1.1               & 1.0            & 1.0        & 1.4                                         & 1.6                                              \\
\bottomrule
\end{tabular}%
}
\caption{Statistics of \ours~decomposed by different collection methods. \ours~exhibits a great deal of topical variety in terms of the number of unique predicates, the number of unique triples, and the vocabulary size.}
\label{tab:dart}
\end{table*}

\subsection{Tree Ontology and Sentence Annotation on Tables}
\label{sec:human_annotation}

\begin{figure*}[t!]
  \centering
  \includegraphics[width=0.85\textwidth]{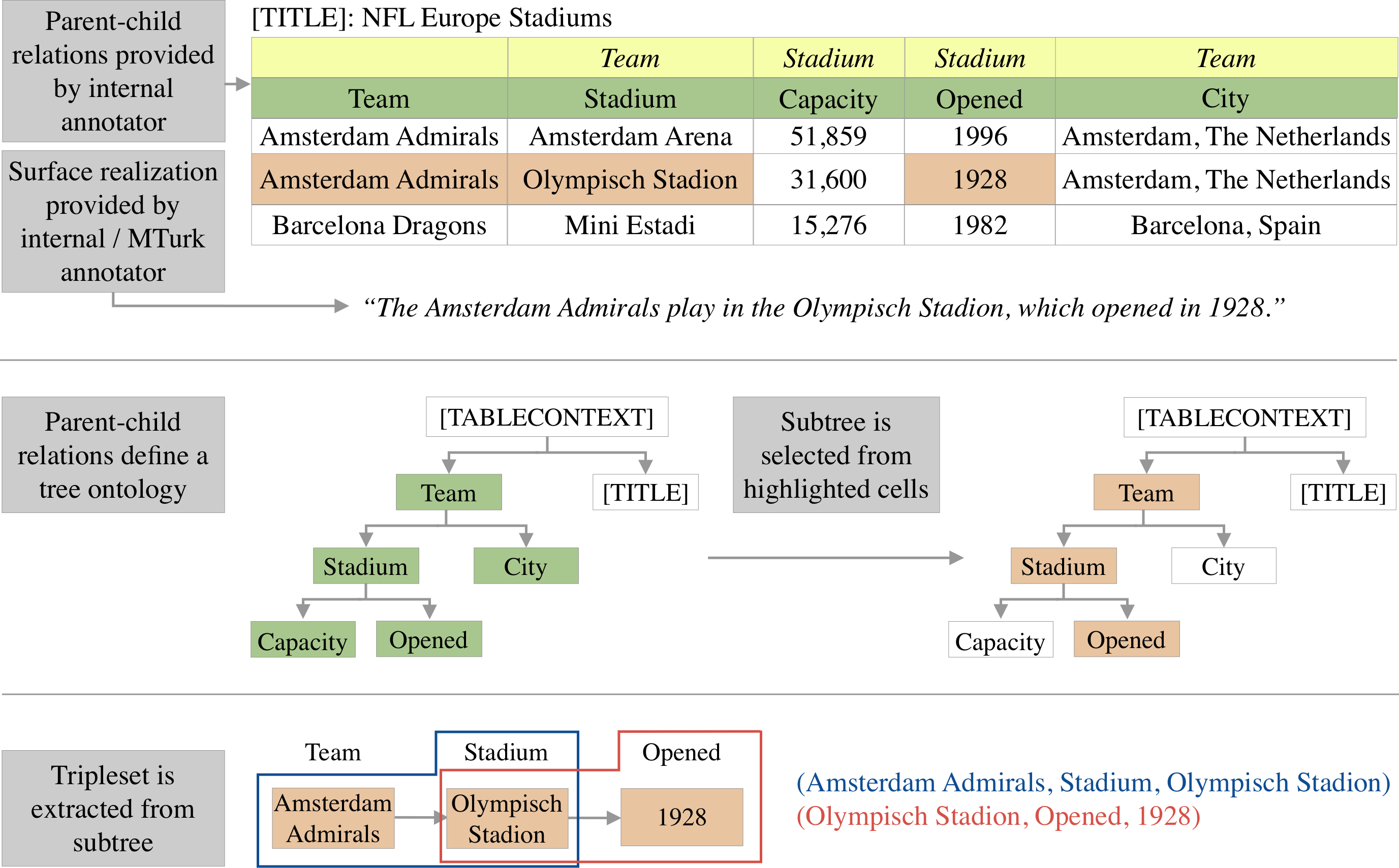}
  \caption{Overview of our human annotation procedure. \textit{Top panel:} We collect the parent-child relations between columns from internal annotators (yellow is parent, green is child). Then, we collect a surface realization of the cells highlighted in orange.  \textit{Middle panel:} We use the provided parent-child relations to construct an ontology tree on the columns, then select the nodes corresponding to the highlighted cells. We gather a connected subtree by collecting all nodes leading up to the highlighted cells' lowest common ancestor. \textit{Bottom panel:} We extract a set of triples from the subtree as shown. This triple-set is paired with the provided realization to form a DART instance.
  }
  \label{fig:human_annotation}
\end{figure*}
Tables are a major source of structured data that contain a wealth of information complementary to text and knowledge graphs.
We aim to collect $\langle$triple-set, sentence$\rangle$ pairs from open-domain Wikipedia tables. However, table schema are flat, making them not directly usable for building subject-predicate-object triples to capture rich relationships in the data. 

As shown in Figure \ref{fig:human_annotation}, we propose a two-stage annotation process that involves two groups of annotators: internal annotators and Amazon Mechanical Turk\footnote{\url{https://www.mturk.com/}} workers. In the first stage, skilled internal annotators specify the parent of every column header to construct a tree-structured ontology for each table. In the second stage, both internal and external annotators provide a sentential description of the highlighted cells in a row that are automatically-chosen based on the ontology.

\paragraph{Tree Ontology Annotation}
For each column in a given table, our internal annotators labeled its ontological parent. In Figure \ref{fig:human_annotation}, for example, the annotator would provide the sequence \{\textsc{null}, \textsc{Team}, \textsc{Stadium}, \textsc{Stadium}, \textsc{Team}\} as the parent of each column --- column \textsc{Team} has no parent, \textsc{Stadium} has parent \textsc{Team}, and so on. In many cases, the relationship between a parent column and its child column can be conceptualized as a "has-a" relationship. For tables that are malformed or have duplicate or missing column names (as shown in Figure \ref{fig:malformed_example} of the Appendix), annotators either changed or added appropriate column names in order to fit these patterns. 
For each table we generate an ontology tree whose root is always [TABLECONTEXT]. This root node either has (1) one child node [TITLE] in the cases where the table title is the subject of entire table, or (2) column header node(s) and a [TITLE] node as children, as shown in Figure \ref{fig:human_annotation}.
This is because in some tables, the table title itself is more appropriate to be the root of the ontology tree (example shown in Figure \ref{fig:title_example} of the Appendix). 
In these cases, annotators assigned the special token [TITLE] as the parent of the relevant column nodes. For other tables, title usually provides important context for understanding the table's rows (example shown in Figure \ref{fig:human_annotation_two} of the Appendix). 
In such cases, [TITLE] is made a child of [TABLECONTEXT] together with the column headers that are appropriate.

We evaluate the quality of the initial tree ontology annotation and made corrections with the following procedure: (1) reject and request corrections from the original annotators if the provided ontology is disconnected or contains a cycle, (2) verify that all column headers appear as a node in the tree. 
For many tables, the determination of an ontology is a subjective process with many "correct" answers - for example, swapping the positions of \textsc{Team} and \textsc{City} in the tree in Figure \ref{fig:human_annotation} produces an equally valid ontology for the referenced table. If there are multiple ways to construct an ontology based on annotators' decisions of attribute relationships among column headers, we manually unify the annotations for similar tables (for examples, tables about athletes in different sports). 
The ontologies exhibit a great deal of structural variety. Relevant statistics are summarized in Table \ref{tab:ontology_data} and Figure \ref{fig:ontology_distribution} of the Appendix.

\paragraph{Connected Component Extraction}
After we annotated the ontology, we automatically choose a subset of cells for a selected table row to form the triple set. Randomly selecting cells leads to poor quality annotation as the selected data could lack a subject, lack cohesion, or would require information not encoded in the ontology to form a coherent sentence. For example, in Figure \ref{fig:human_annotation}, if only two nodes \textsc{City} and \textsc{Capacity} were highlighted then a coherent sentence cannot be produced as there is no direct logical relationship (functional dependency) between them. To solve these issues, instead of randomly selecting cells in a row, we extract connected components from the ontology. 

The extracted components have two controllable properties: size and shape. To create variation in size, we randomly sampled between $[2, 5]$. The shape is determined by two numbers: the number of sibling node pairs and parent-child node pairs. Increasing the number of sibling node pairs creates a wider tree, while increasing the latter creates a deeper tree. We created a sliding scale between width and depth using an expansion parameter, $p$. We recursively visit a node if it has children with probability $p$ and otherwise move to a sibling if it exists. If $p$ = 1, the search becomes a DFS and if $p$ = 0, it becomes BFS. We found that randomly selecting $p$ from 0.5 to 0.7 created a reasonable variation in extracted component shapes. This ensures the balance between breadth and depth of ontology coverage of the selected cells, therefore ensuring the quality of the sentence annotation.

\paragraph{Sentence Annotation}
Given the table, title, and connected highlighted cells of a row, annotators were asked to write a description of the highlighted cells. We encouraged the annotators to use diverse vocabulary and syntactic structures. 
To ensure quality, internal annotators reviewed every crowd sourced sentence for correctness.
They either rewrote or discarded the sentences that were nonsensical or incorrect. In some cases, they also changed cell highlighting patterns to match the sentence provided.

\paragraph{Build Tripleset-Sentence Pairs}
Finally, we convert the highlighted cells to triplesets. For a row $R$, we start with the table's column ontology $T$. We first place the cell values in $R$ in their corresponding slots in $T$, e.g. in Figure \ref{fig:human_annotation} we fill \textsc{Team} with "Amsterdam Admirals". We then check that the nodes of $T$ corresponding to the highlighted cells in $R$ form a connected subtree. If not, we walk up the tree and highlight each traversed node up until the lowest common ancestor of the highlighted nodes (inclusive) to form a connected subtree. For each node $N$ in the tree except the root node, we can extract the triple ($\mathrm{parent}\left(N\right)$, $\mathrm{title}\left(N\right)$, $N$). For example, since \textsc{Stadium} is highlighted in Figure~\ref{fig:human_annotation}, we extract the triple (Amsterdam Admirals, \textsc{Stadium}, Olympisch Stadion).
A small number of triple-sets contained more than 10 triples. We discarded these because their associated surface realizations were of poor quality. The numbers of tripleset-sentence pairs annotated by different annotators are shown in Table \ref{tab:dart}.

\subsection{Automatically Converting Questions to Declarative Sentences}
\label{sec:declarative}
High quality natural language questions in open domain semantic parsing datasets such as WikiSQL and QA2D techniques found in automatically constructing NLI datasets \cite{demszky2018transforming} present themselves as an attractive opportunity to semi-automatically construct an abundance of declarative sentences and align to table cells. We leveraged rule-based QA2D technique\footnote{We use the rule-based model from \url{https://github.com/kelvinguu/qanli} \cite{demszky2018transforming}. The neural model code is not released.} together with manual screening to combine WikiSQL questions and SQL-retrieved-answers into declarative sentences and manually filtered out bad sentences.

We only execute SQL queries without aggregate commands\footnote{ \textsc{MAX}, \textsc{MIN}, \textsc{COUNT}, \textsc{SUM}, \textsc{AVG}, \textsc{JOIN}, \textsc{INTERSECT}, \textsc{UNION}, \textsc{GROUP BY}, \textsc{ORDER BY}.} to retrieve  answers corresponding to questions answerable by single rows.
An example of such conversion is as follows: \\\\
Question: \textit{In which year did Greece hold its last Summer Olympics?}\\
Answer: \textit{2004}\\
Declarative Sentence: \textit{Greece held its last Summer Olympics in 2004.}\\
\\\noindent
Alignment with table cells is done at two stages. We first align sentences with corresponding rows by changing SQL commands to \textsc{SELECT *} and use string matching to obtain columns and column headers relevant to the answer and \textsc{WHERE} condition. After manually filtering out bad sentences, bad alignments, or tables without ontology annotations, we were able to get 4,204 sentences.
Finally, the corresponding table cells are then converted into triples in the same way as we described in Section~\ref{sec:human_annotation}. 

Examples of produced declarative sentences can be found in Figure~\ref{fig:WSQL_Declarative_Example} of the Appendix.

\subsection{Incorporating Existing Datasets}
\label{sec:existing}
Since they provide a large amount of strictly aligned data-text pairs with high quality sentences, we incorporate the following existing datasets in the same $\langle$triple-set, sentence$\rangle$ pair format with some modifications.

\paragraph{WebNLG 2017} An instance of the WebNLG dataset contains a set of triples extracted from DBpedia and the target text written by human. We include the WebNLG 2017 dataset\footnote{\url{https://gitlab.com/shimorina/webnlg-dataset/-/tree/master/webnlg_challenge_2017}} consisting of 27731 triple-set sentence pairs with up to 7 RDF triples in a triple set covering 15 domains. 

\paragraph{Cleaned E2E} The original E2E dataset includes dialogue act meaning representations (MR) and natural language references in the restaurant domain. 
Later, \citet{dusek2019semantic} provide Cleaned E2E\footnote{\url{https://github.com/tuetschek/e2e-cleaning}} by automatically fixing the dialogue acts to account for omissions and hallucinations in the text. We incorporate Cleaned E2E because of its strict alignment between the meaning representation and the text.
To convert the MR to a triple-set, we take the \textsc{name} slot (present in almost all the MRs) as the subject. For example, the MR \textsc{(name[Alimentum], area[city centre], familyFriendly[no])} is converted to the triple-set \textsc{\{(Alimentum, area, city centre), (Alimentum, familyFriendly, no)\}}.
We drop MRs which do not contain the \textsc{name} slot.

\subsection{Predicate Unification}
\label{sec:unification}
We canonicalized the predicates in our triple sets such that those of the same meaning are also represented the same.
We manually constructed a predicate mapping table to achieve this.
As an example, our predicate mapping maps "Hometown," "Home Town," and "Home Town/City" to the unified predicate "HOMETOWN."

After unifying predicates, we evaluated the diversity of~\ours by counting the number of unique predicates in its partitions. 
As shown in Table \ref{tab:dart}, we see that the Wikipedia partition of \ours contains much more unique predicates than the WebNLG and Cleaned E2E partitions combined, despite having smaller number of $\langle$triple-set, sentence$\rangle$ pairs. This contributes significantly to the domain diversity of \ours. In addition, we can see that \ours~exhibits a great deal of topical variety in terms of number of unique triples and vocabulary size.

\subsection{Dataset Split}
\label{sec:dataset_split}

For WebNLG 2017 and Cleaned E2E, we use their original data splits.
For our annotation on WikiTableQuestions and WikiSQL, random splitting will make train, dev, and test splits contain similar tables and similar $\langle$triple-set, sentence$\rangle$ examples.
Therefore, to increase the generalization challenge, we compare the table title and the table header to find similar tables, and make sure the model is evaluated on test split tables that are least similar to those used for training.
We first sample some tables as a seed test set, and then compute Jaccard similarity\footnote{\url{https://en.wikipedia.org/wiki/Jaccard_index}} with remaining tables based on the titles and the headers.
If a table has a Jaccard similarity greater than 0.5 with any of the tables in the test set, we add it into the test set.
A similar process is repeated to create the dev set, and the remaining tables form the training set.
This results in 62,659/6,980/12,552 sentences in the train/dev/test sets, respectively.

\section{Experimental Results}
We conduct experiments on DART and the WebNLG 2017 dataset, with an ablation study on WebNLG to show the benefits of using DART for data augmentation.

\begin{table*}[ht!]
\centering
\resizebox{\textwidth}{!}{%
\begin{tabular}{lcccccccc}
\toprule
                            & BLEU $\uparrow$  & METEOR $\uparrow$ & TER $\downarrow$ & MoverScore $\uparrow$ & BERTScore(F1) $\uparrow$ & BLEURT $\uparrow$ & PARENT $\uparrow$ \\ \hline
LSTM with Attention   & 29.66 & 0.27 & 0.63 & 0.31 & 0.90 & -0.13 & 0.35 \\
End-to-End Transformer      & 27.24 & 0.25 & 0.65 & 0.25 & 0.89 & -0.29 & 0.28 \\
BART-base                   & 47.11 & 0.38 & 0.46 & 0.51 & 0.95 & 0.37 & 0.55\\
BART-large                  & 48.56 & 0.39 & 0.45 & 0.52 & 0.95 & 0.41 & 0.57\\
T5-small                    & 47.69 & 0.39 & 0.46 & 0.52 & 0.95 & 0.40 & 0.56\\
T5-base                     & 49.21 & 0.40 & 0.44 & 0.53 & 0.95 & 0.43 & 0.57\\
T5-large                    & \bf{50.66} & \bf{0.40} & \bf{0.43} & \bf{0.54} & \bf{0.95} & \bf{0.44} & \bf{0.58} \\
\bottomrule
\end{tabular}%
}
\caption{Model results on the test set of \ours~ $\uparrow$: Higher is better. $\downarrow$: Lower is better.}
\label{tab:dart_result}
\end{table*}

\begin{table*}[ht!]
\centering
\resizebox{\textwidth}{!}{%
\begin{tabular}{lccccccccc}
\toprule
             & \multicolumn{3}{c}{BLEU $\uparrow$}  & \multicolumn{3}{c}{METEOR $\uparrow$} & \multicolumn{3}{c}{TER $\downarrow$}  \\ \hline
             & SEEN  & UNSEEN & ALL & SEEN  & UNSEEN & ALL & SEEN  & UNSEEN & ALL \\ \hline
Pipeline Transformer$^{\dagger}$  \cite{ferreira2019neural} & 56.28 & 23.04 & 42.41 & 0.42 & 0.21 & 0.32 & 0.39 & 0.63 & 0.50 \\
Pipeline GRU$^{\dagger}$ \cite{ferreira2019neural}   & 56.09 & 25.12 & 42.73 & 0.42 & 0.22 & 0.33 & 0.39 & 0.64 & 0.51 \\
MELBOURNE    \cite{gardent2017webnlg}    & 54.52 & 33.27 & 45.13 & 0.41 & 0.33 & 0.37 & 0.40 & 0.55 & 0.47 \\
BestPlan $^{\dagger}$ \cite{moryossef2019step}  & 53.30 & 34.41 & 47.24 & 0.44 & 0.34 & 0.39 & 0.47 & 0.56 & 0.51 \\
DualEnc      \cite{zhao2020bridging}   & 63.45 & 36.73 & 51.42 & 0.46 & 0.37 & 0.41 & 0.34 & 0.55 & 0.44 \\
PlanEnc      \cite{zhao2020bridging}   & 64.42 & 38.23 & 52.78 & 0.45 & 0.37 & 0.41 & 0.33 & 0.53 & 0.42 \\
\midrule 
\citet{ribeiro2020investigating} \\
\quad BART-base $^{\ddagger}$ & 63.02 & 41.74 & 53.36 & 0.45 & 0.35 & 0.40 & 0.33 & 0.52 & 0.42 \\
\quad BART-large $^{\ddagger}$   & 63.71 & 44.17 & 54.95 & 0.46 & 0.39 & 0.42 & 0.33 & 0.51 & 0.41 \\
\quad T5-small $^{\ddagger}$     & 65.30 & 45.58 & 56.57 & 0.46 & 0.39 & 0.43 & 0.32 & 0.49 & 0.40 \\
\quad T5-base $^{\ddagger}$      & 64.89 & 52.86 & 59.44 & 0.46 & 0.42 & 0.44 & 0.33 & 0.42 & 0.37 \\
\quad T5-large $^{\ddagger}$     & 64.89 & 54.01 & 59.95 & 0.46 & 0.43 & 0.44 & 0.34 & 0.41 & 0.37 \\
\midrule
\textbf{+ \ours} \\
\quad BART-base & 62.36 & 46.21 & 55.14 & 0.44 & 0.37 & 0.41 & 0.34 & 0.45 & 0.39 \\
\quad BART-large & 64.51 & 50.20 & 58.06 & 0.46 & 0.40 & 0.43 & 0.32 & 0.44 & 0.38 \\
\quad T5-small & 65.05 & 47.81 & 57.32 & 0.46 & 0.40 & 0.43 & 0.33 & 0.46 & 0.39 \\
\quad T5-base  & 65.42 & 50.71 & 58.80 & 0.46 & 0.41 & 0.44 & 0.32 & 0.43 & 0.37 \\ 
\quad T5-large & \bf{65.82} & \bf{56.01} & \bf{61.44} & \bf{0.46} & \bf{0.43} & \bf{0.45} & \bf{0.32} & \bf{0.38} & \bf{0.35} \\

\bottomrule
\end{tabular}%
}
\caption{The WebNLG 2017 results on the test set.
$^{\dagger}$: We report results from \citet{zhao2020bridging} who use the evaluation scripts that are strictly the same as the official challenge.
$^{\ddagger}$: We report results calculated with the model outputs on the WebNLG 2017 testset released by \citet{ribeiro2020investigating}.
}
\label{tab:webnlg}
\end{table*}

\subsection{Models}
We investigate several state-of-the-art Data-to-Text generation models. We report results of the following models on DART-testset: (1) Bidirectional-LSTM with attention, for which we use 2-layer bi-LSTM for encoder, with 300 dimensional word embeddings (without using pretrained word vectors), 512 hidden units and 0.3 dropout rate for the decoder. (2) Transformer \cite{vaswani2017attention}, previously used by \citet{ferreira2019neural} on the WebNLG dataset. The input is formed by linearizing the unordered triple set. (3) BART \cite{lewis2020bart}, for which we report results of both BART-base and BART-large. (4) T5 \cite{raffel2020exploring}: we add the same prefix "translate Graph to English:" to the input, as it is used in \citet{ribeiro2020investigating}. We report results of T5-small, T5-base and T5-large models. For both BART and T5 models, we use implementations of \citet{ribeiro2020investigating}, with same hyperparameter setting.

\subsection{Evaluation Metrics}
We use a variety of automatic metrics and human evaluation (Section \ref{sec:human_eval}) to evaluate the quality of the generated text.
We report BLEU, METEOR, and TER which are used in the official WebNLG challenge. However, these measures have limitations in considering the semantic meanings of words or phrases \cite{novikova2017we}, therefore we also report MoverScore \cite{zhao2019moverscore}, BERTScore \cite{zhang2020bertscore}, and BLEURT \cite{sellam2020bleurt} that incorporate semantics rather than surface forms using contextual embeddings.
Furthermore, we include PARENT \cite{dhingra2019handling} which explicitly aligns n-grams from the reference and generated text to the data contents.

\begin{table*}[ht!]
\centering 
\resizebox{.9\textwidth}{!}{%
\begin{tabular}{cccccc}
\toprule
\textbf{Tripleset source} & \textbf{Sentence source} & \textbf{\% fluent} & \textbf{\% faithful} & \textbf{\begin{tabular}[c]{@{}c@{}}\% (fluent+\\ mostly fluent)\end{tabular}} & \textbf{\begin{tabular}[c]{@{}c@{}}\% (faithful+\\ mostly faithful)\end{tabular}} \\ \midrule
\multirow{3}{*}{\textbf{\begin{tabular}[c]{@{}c@{}}WikiTableQuestions (\S~\ref{sec:human_annotation})\end{tabular}}} 
 & human-written reference & 75\% & 81\% & 96\% & 99\% \\
 & BART-base & 74\% & 57\% & 93\% & 84\% \\
 & T5-base   & 72\% & 54\% & 94\% & 76\% \\ \midrule
\multirow{3}{*}{\textbf{\begin{tabular}[c]{@{}c@{}}WikiSQL (\S~\ref{sec:declarative})\end{tabular}}} 
 & auto-generated reference & 59\% & 56\% & 87\% & 88\% \\
 & BART-base & 66\% & 51\% & 92\% & 83\% \\
 & T5-base   & 75\% & 65\% & 97\% & 90\% \\ \bottomrule
\end{tabular}%
}
\caption{Human evaluation over references and model outputs.}
\label{tab:human_eval}
\end{table*}

\subsection{Results}
\paragraph{\ours}
Our experimental results on \ours~are summarized in Table \ref{tab:dart_result}. 
The T5-large model has the highest performance among all models with a BLEU score of 50.66. We attribute this to T5's generalization and transfer learning ability due to pretraining on multi-tasks. 
We can see that in general, pretrained models outperform others by a large margin, and increasing the model size seems to further boost the performance on DART. However, language models such as BART and T5 are pretrained by reconstructing text and, as a result, we found that their output on \ours~often contains hallucinated words \cite{parikh2020totto,harkous2020have,reiter2020openai}, as shown in Figure \ref{fig:hallucination_examples}.
In addition, while the pretrained model shows better text generation quality due to its generalization ability from pretraining, it does not fully capture the hierarchical ontology nature of the triple sets in their linearized input, therefore making DART more challenging. We suspect that models that are better at exploiting the ontology structure preserved in the input tripleset will achieve better performance on DART. 

\paragraph{WebNLG}
Furthermore, we investigate if \ours~can improve pretrained models' performance on other Data-to-Text generation tasks.
To this end, we finetune the baseline transformer model, BART-[base, large] and T5-[small, base, large] on the WebNLG 2017 dataset, and augment the training by adding instances in the \ours~training set. The experimental results can be found in Table \ref{tab:webnlg}. We report performances of some competitive models that are not pretrained, as well as the state-of-the-art performances of pretrained models on the WebNLG 2017 dataset by \citet{ribeiro2020investigating}. On the bottom panel, we include results of experiments augmented with DART instances whose triplesets are generated with table ontology annotation, paired with human written sentences.
We are able to achieve new state-of-the-art results on all WebNLG 2017 test set splits (seen, unseen and all) by finetuning T5-large on \ours. We observe that using \ours~for data augmentation consistently improves the performance across all models, including the baseline transformer model that is not pretrained. Furthermore, we observe that more improvement is shown on unseen split of the test set, due to \ours's open-domain nature. See Figure \ref{fig:data_augmentation_output_examples} of the Appendix for example model outputs aligned with their human references. 

\subsection{Ablation Study}
\label{sec:ablation}
We also conduct an ablation study on the WebNLG dataset to investigate what part of DART contributes most to improving the Data-to-Text tasks in general. We report results of the study in Table \ref{tab:ablation_study} of the Appendix. We divide DART into 4 partitions, where declarative sentence (auto-generated) partition and human annotated sentence partition contain instances whose triplesets are extracted from Wikipedia tables based on ontology.  E2E partition contains instances converted from the E2E dataset, and WebNLG partition keeps the original data format. In general, we observe that adding DART instances that contain human written sentences brings most improvement, especially on unseen split. While adding E2E partition boosts the scores on seen test split and deteriorates the performance on unseen test split. This trend is consistent across all models. Comparing results of declarative sentence partition and human written sentence partition, we see that for most of the models, DART instances with human written sentences have better quality as it brings more improvement to the task.

\section{Human Evaluation}
\label{sec:human_eval}

In Table \ref{tab:human_eval}, we perform human evaluation on \ours~based on two criteria: (1) \emph{fluency} if a sentence is natural and grammatical, and (2) semantic \emph{faithfulness} if a sentence is supported by the input triples.
We defined three levels of fluency: fluent, mostly fluent, and not fluent, and the same for semantic faithfulness. 
We ask 5 internal annotators to evaluate on 100 triplesets sampled from declarative sentence partition and another 100 triplesets sampled from human written sentence partition.
Each tripleset is paired with 3 sentences, one of them is the reference sentence, and the other two are outputs of BART-base and T5-base models.

The results in Table \ref{tab:human_eval} attest to the high quality of our annotations since the human written references achieve highest fluency and faithfulness comparing to outputs of two strong baseline models. The evaluation on faithfulness also demonstrates that there is a considerable gap between the DART reference and the outputs of the state-of-the-art pretrained model, showing that there is a large room for improvement. We also noticed that the auto-generated declarative sentences are not as fluent or faithful as the model outputs because they are generated with a rule-based system.
However, we decided to release this partition, along with other partitions of \ours~because it demonstrates an economic way to obtain large amounts of DART instances and it also shows benefits for generalization due to the diverse topics it contains.

\section{Related Work}
\paragraph{Data-to-Text}
Data-to-Text generation aims to produce natural language output from structured input. Applications include generating sports commentaries~\citep{chen2008learning,wiseman2017challenges},
weather forecasts~\citep{liang2009learning,konstas2012unsupervised},
biographical texts~\citep{lebret2016neural, liu2018table},
knowledge-base descriptions~\citep{gardent2017webnlg},
dialogue response generation~\citep{wen2015semantically,wen2016conditional}, 
and commonsense reasoning~\citep{lin2019comgen}.
Yet, most existing datasets are restricted to specific domains and applications.
In contrast, a major source of \ours~is from Wikipedia tables covering various domains and topics.
\paragraph{Representation of Data} 
The input of the Data-to-Text datasets take different formats, including slot-value pairs, Abstract Meaning Representation (AMR)~\cite{song2017amr,ribeiro2019enhancing},
Minimal Recursion Semantics (MRS) \cite{hajdik2019neural}, 
Resource Description Framework (RDF triples)~\cite{gardent2017webnlg}, and logic forms \cite{chen2020logic2text}. There are also studies of converting tabular data to RDF triples in the Semantic Web community \cite{Kellogg:15:GRT}.
Recently, some open-domain table-to-text datasets have been proposed including WikiTableText \cite{bao2018table}, LogicNLP \cite{chen2020logical}, and ToTTo \cite{parikh2020totto}, whose inputs are rows or entire tables. In ToTTo, highlighted cells are also provided as input, and the authors found using only highlighted cells with flat row and column headers led to higher performance than using the entire table. 

In contrast, \ours~is constructed by first annotating the tree-structured table ontology that encodes the semantic dependencies among table headers, and we could flexibly incorporate additional contexts such as the table title to the ontology tree. We then use an automatic procedure to extract connected components from the tree to form the input of a DART instance. Our annotation framework not only provides a flexible way of incorporating any contexts to the representation of tables, but also encodes hierarchical relationships among table headers and contexts, ensuring the extracted triples are logically consistent and can be described in text without loss of information.

\paragraph{Model}
Traditional Data-to-Text models break the generation progress into different stages such as signal analysis, data interpretation, document planning, microplanning, and realization \cite{reiter2000building,reiter2007architecture}.
Recently, neural encoder-decoder models based on attention and copy mechanisms have shown promising results~\citep{gehrmann2018end,puduppully2018data,puduppully2019data,ferreira2019neural}.
Furthermore, recent progress on pretrained models such as GPT-2 \cite{radford2018improving}, BART \cite{lewis2020bart} and T5 \cite{raffel2020exploring} has shown effective results for text generation tasks on machine translation, summarization, and conversation response generation.
\citet{chen2020few,peng2020few,kale2020text} also finetune pretrained models on Data-to-Text tasks.

\section{Conclusion}
In this paper, we introduce \ours, an open-domain corpus for structured data record to text generation.
\ours's ontology-preserving representation of data inputs differentiates itself from other open-domain Data-to-Text corpora.
We found that \ours~introduces new challenges to several state-of-the-art Data-to-Text models due to its open-domain nature and its ontology structure of the semantic triple input. Furthermore, we found that using it for data augmentation improves other Data-to-Text tasks.
For future work, we will explore more controlled, high-fidelity generation that better incorporates the ontology hierarchy of data.

\newpage
\section{Ethics Statement}
Our dataset is constructed by accumulating and processing resources from various existing datasets that are open to the public. In addition, we collect annotations on structure of tabular data and human written sentences that describe data records. 

The existing resources that we utilize mainly consist of (1) tabular data from Wikipedia, (2) information of restaurants presented with dialogue-act meaning representation and its textual description (E2E), and (3) information of various entities and their relationship that are in 15 different categories of DBPedia, which is a knowledge base built on contents created in various Wikimedia projects (WebNLG). It is possible that there are biases in these resources, either in the tabular data or the textual description written by humans. 

For additional annotations we collected, we have two groups of annotators participating: internal annotators who are the authors of this work, and external annotators recruited from the Amazon Mechanical Turk platform.
On MTurk, we use a pay rate of \$15 per hour approximately based on our estimation of the time it takes to complete our annotation tasks. In total, it took 125 hours to complete all tasks on the Amazon Mechanical Turk platform.
There are three annotation tasks: (1) Annotators are asked to specify ontological structure of the table by indicating relationship between table column headers, (2) Annotators are asked to write descriptions that are fluent and semantically faithful to the data records presented to them, and (3) Annotators are asked to evaluate sentences that are either references or model generated outputs. We acknowledge that it is also possible to have biases in the sentences written by the annotators, or in the data records that are presented to them.

We conducted experiments on our own dataset and the WebNLG dataset using BART and T5, two large-scale pretrained models. Both models are trained on large amounts of textual data such as news, books, and web text, which may contain any kinds of biases. As a result, it is possible to insert those biases into the models. 

In total, we conducted 43 experiments: 7 on DART and 36 for our ablation study on the WebNLG dataset. We use a single NVIDIA V100 GPU for all experiments and each experiment took from 5 to 40 hours depending on the model size. 

\section*{Acknowledgement}
The authors would like to thank the anonymous reviewers for their discussion and feedback.

\bibliography{anthology,custom}
\bibliographystyle{acl_natbib}

\clearpage
\newpage
\appendix
\onecolumn
\section*{Appendix}
\label{sec:appendix}

The Appendix contains the following contents:

\begin{itemize}
  \item Results of the ablation study on WebNLG 2017 testset.
  \item Statistics of the table ontology annotations.
  \item Examples of tables that help illustrate DART's annotation procedure.
  \item Examples of model outputs.
\end{itemize}


\begin{table*}[ht!]

\resizebox{\linewidth}{!}{%
\begin{tabular}{clccccccccc}
\toprule
\multirow{2}{*}{\textbf{Model}}                                                          & \multicolumn{1}{c}{\multirow{2}{*}{\textbf{Experiment}}} & \multicolumn{3}{c}{\textbf{BLEU $\uparrow$}}                 & \multicolumn{3}{c}{\textbf{METEOR $\uparrow$}}             & \multicolumn{3}{c}{\textbf{TER $\downarrow$}}                \\
                                                                                         & \multicolumn{1}{c}{}                                     & \textbf{SEEN}  & \textbf{UNSEEN} & \textbf{ALL}   & \textbf{SEEN} & \textbf{UNSEEN} & \textbf{ALL}  & \textbf{SEEN} & \textbf{UNSEEN} & \textbf{ALL}  \\ \midrule
\multirow{6}{*}{\textbf{\begin{tabular}[c]{@{}c@{}}Baseline\\ Transformer\end{tabular}}} & {[}1{]} webnlg                                           & 49.81          & 5.51            & 31.81          & 0.39          & 0.09            & 0.24          &     0.47          &         0.86        &       0.64        \\
                                                                                         & {[}2{]} webnlg+dart\_decl\_sents                         & 52.31          & \textbf{8.96}            & \textbf{39.98}          & 0.40          & 0.07            & 0.25          &    0.45           &         0.79        &   0.60            \\
                                                                                         & {[}3{]} webnlg+dart\_human\_annotated                    & 53.68          & 7.02            & 36.36          & 0.40          & \textbf{0.09}            & \textbf{0.26}          &     0.43          &         \textbf{0.79}        &   \textbf{0.59}            \\
                                                                                         & {[}4{]} webnlg+dart\_ontology                            & 53.40          & 8.54            & 38.51          & \textbf{0.41}          & 0.08            & 0.26          &   0.44            &       0.80          &       0.60        \\
                                                                                         & {[}5{]} webnlg+dart\_e2e                                 & 51.76          & 5.92            & 32.36          & 0.40          & 0.09            & 0.25          &    0.45           &         0.86        &       0.63        \\
                                                                                         & {[}6{]} webnlg+dart\_full                                & \textbf{54.99}          & 8.64            & 39.11          & 0.40          & 0.08            & 0.25          &    \textbf{0.42}           &         0.81        &   0.60            \\ \midrule
\multirow{6}{*}{\textbf{BART-base}}                                                      & {[}1{]} webnlg                                           & 63.02          & 41.74           & 53.36          & 0.45          & 0.35            & 0.40          & 0.33          & 0.52            & 0.42          \\
                                                                                         & {[}2{]} webnlg+dart\_decl\_sents                         & 62.71          & 42.51           & 53.64          & 0.45          & 0.36            & 0.40          & 0.34          & 0.51            & 0.41          \\
                                                                                         & {[}3{]} webnlg+dart\_human\_annotated                    & 62.36          & 46.21           & 55.14          & 0.44          & 0.37            & 0.41          & 0.34          & 0.45            & 0.39          \\
                                                                                         & {[}4{]} webnlg+dart\_ontology                            & 62.62          & \textbf{46.74}  & \textbf{55.54} & 0.44          & \textbf{0.38}   & \textbf{0.41} & 0.34          & \textbf{0.45}   & \textbf{0.39} \\
                                                                                         & {[}5{]} webnlg+dart\_e2e                                 & \textbf{64.00} & 35.07           & 51.17          & \textbf{0.45} & 0.33            & 0.40          & \textbf{0.33} & 0.61            & 0.46          \\
                                                                                         & {[}6{]} webnlg+dart\_full                                & 63.66          & 45.48           & 55.52          & 0.45          & 0.37            & 0.41          & 0.33          & 0.47            & 0.40          \\ \midrule
\multirow{6}{*}{\textbf{BART-large}}                                                     & {[}1{]} webnlg                                           & 63.71          & 44.17           & 54.95          & 0.46          & 0.39            & 0.42          & 0.33          & 0.51            & 0.41          \\
                                                                                         & {[}2{]} webnlg+dart\_decl\_sents                         & 65.18          & 46.79           & 56.79          & 0.46          & 0.39            & 0.42          & 0.32          & 0.48            & 0.40          \\
                                                                                         & {[}3{]} webnlg+dart\_human\_annotated                    & 64.51          & \textbf{50.20}  & \textbf{58.06} & 0.46          & \textbf{0.40}   & \textbf{0.43} & 0.32          & \textbf{0.44}   & \textbf{0.38} \\
                                                                                         & {[}4{]} webnlg+dart\_ontology                            & 64.19          & 49.62           & 57.65          & 0.46          & 0.39            & 0.43          & 0.33          & 0.45            & 0.38          \\
                                                                                         & {[}5{]} webnlg+dart\_e2e                                 & 65.06          & 30.17           & 48.24          & 0.46          & 0.33            & 0.40          & 0.32          & 0.69            & 0.49          \\
                                                                                         & {[}6{]} webnlg+dart\_full                                & \textbf{65.24} & 47.96           & 57.44          & \textbf{0.46} & 0.39            & 0.43          & \textbf{0.32} & 0.46            & 0.39          \\ \midrule
\multirow{6}{*}{\textbf{T5-small}}                                                       & {[}1{]} webnlg                                           & 65.30          & 45.58           & 56.57          & 0.46          & 0.39            & 0.43          & 0.32          & 0.49            & 0.40          \\
                                                                                         & {[}2{]} webnlg+dart\_decl\_sents                         & 64.18          & 46.61           & 56.27          & 0.46          & 0.39            & 0.43          & 0.33          & 0.48            & 0.40          \\
                                                                                         & {[}3{]} webnlg+dart\_human\_annotated                    & 65.05          & \textbf{47.81}  & \textbf{57.32} & 0.46          & \textbf{0.40}   & \textbf{0.43} & 0.33          & \textbf{0.46}   & \textbf{0.39} \\
                                                                                         & {[}4{]} webnlg+dart\_ontology                            & 65.17          & 47.49           & 57.24          & 0.46          & 0.39            & 0.43          & 0.32          & 0.47            & 0.39          \\
                                                                                         & {[}5{]} webnlg+dart\_e2e                                 & \textbf{65.56} & 41.28           & 54.56          & \textbf{0.46} & 0.38            & 0.42          & \textbf{0.32} & 0.54            & 0.42          \\
                                                                                         & {[}6{]} webnlg+dart\_full                                & 64.70          & 47.56           & 57.01          & 0.46          & 0.39            & 0.43          & 0.33          & 0.47            & 0.39          \\ \midrule
\multirow{6}{*}{\textbf{T5-base}}                                                        & {[}1{]} webnlg                                           & 64.89          & \textbf{52.86}  & 59.44          & 0.46          & \textbf{0.42}   & 0.44          & 0.33          & \textbf{0.42}   & 0.37          \\
                                                                                         & {[}2{]} webnlg+dart\_decl\_sents                         & 65.44          & 50.80           & 58.81          & 0.46          & 0.41            & 0.44          & 0.32          & 0.43            & 0.37          \\
                                                                                         & {[}3{]} webnlg+dart\_human\_annotated                    & 65.42          & 50.71           & 58.80          & 0.46          & 0.41            & 0.44          & 0.32          & 0.43            & 0.37          \\
                                                                                         & {[}4{]} webnlg+dart\_ontology                            & 65.17          & 51.49           & 59.04          & 0.46          & 0.41            & 0.44          & 0.33          & 0.43            & 0.37          \\
                                                                                         & {[}5{]} webnlg+dart\_e2e                                 & 65.11          & 49.64           & 58.19          & 0.46          & 0.41            & 0.44          & 0.33          & 0.46            & 0.39          \\
                                                                                         & {[}6{]} webnlg+dart\_full                                & \textbf{65.99} & 51.68           & \textbf{59.50} & \textbf{0.46} & 0.42            & \textbf{0.44} & \textbf{0.32} & 0.43            & \textbf{0.37} \\ \midrule
\multirow{6}{*}{\textbf{T5-large}}                                                       & {[}1{]} webnlg                                           & 64.89          & 54.01           & 59.95          & 0.46          & 0.43            & 0.44          & 0.34          & 0.41            & 0.37          \\
                                                                                         & {[}2{]} webnlg+dart\_decl\_sents                         & 65.97          & 53.00           & 60.12          & 0.46          & 0.42            & 0.44          & 0.32          & 0.41            & 0.36          \\
                                                                                         & {[}3{]} webnlg+dart\_human\_annotated                    & 65.82          & \textbf{56.01}  & \textbf{61.44} & 0.46          & \textbf{0.43}   & \textbf{0.45} & 0.32          & \textbf{0.38}   & \textbf{0.35} \\
                                                                                         & {[}4{]} webnlg+dart\_ontology                            & 65.53          & 55.20           & 60.90          & 0.46          & 0.42            & 0.44          & 0.32          & 0.38            & 0.35          \\
                                                                                         & {[}5{]} webnlg+dart\_e2e                                 & \textbf{66.27} & 54.13           & 60.76          & \textbf{0.46} & 0.43            & 0.45          & \textbf{0.32} & 0.41            & 0.36          \\
                                                                                         & {[}6{]} webnlg+dart\_full                                & 65.78          & 54.35           & 60.64          & 0.46          & 0.42            & 0.44          & 0.32          & 0.39            & 0.35          \\ \bottomrule
\end{tabular}%
}
\caption{Results of ablation study on WebNLG 2017 testset. \textbf{dart\_decl\_sents} refers to DART partition that contains auto-generated declarative sentences mentioned in Section \ref{sec:declarative}, \textbf{dart\_human\_annotated} refers to partition that contains human written sentences mentioned in Section \ref{sec:human_annotation}, \textbf{dart\_ontology} is the combination of \textbf{dart\_decl\_sents} and \textbf{dart\_human\_annotated}, and \textbf{dart\_e2e} refers to DART partition containing instances extracted from E2E dataset, the process of which is mentioned in Section \ref{sec:existing}. Note that \textbf{dart\_full} is the combination of \textbf{dart\_ontology} and \textbf{dart\_e2e}.}
\label{tab:ablation_study}
\end{table*}


\begin{table*}[ht!]
\centering
\resizebox{.8\textwidth}{!}{%
\begin{tabular}{lcccccc}
\toprule
                                  & Tables & \shortstack{Ontology depth \\ (min, med, max)} & \shortstack{Nodes in ontology \\ (min, med, max)} & \shortstack{ Branching factor \\ (mean)} \\ \hline
WikiTableQuestions &  2060 & 1, 1, 4 & 2, 6, 25 & 4.0 \\
WikiSQL            &  3563 & 1, 1, 4 & 3, 7, 25 & 5.1\\
\bottomrule
\end{tabular}%
}
\caption{Properties of the ontology in the WikiTableQuestions and WikiSQL samples in \ours. Branching factor refers to the average number of children across all non-leaf nodes in a table's ontology.}
\label{tab:ontology_data}
\end{table*}


\begin{figure*}[t!]
  \centering
  \includegraphics[width=0.45\textwidth]{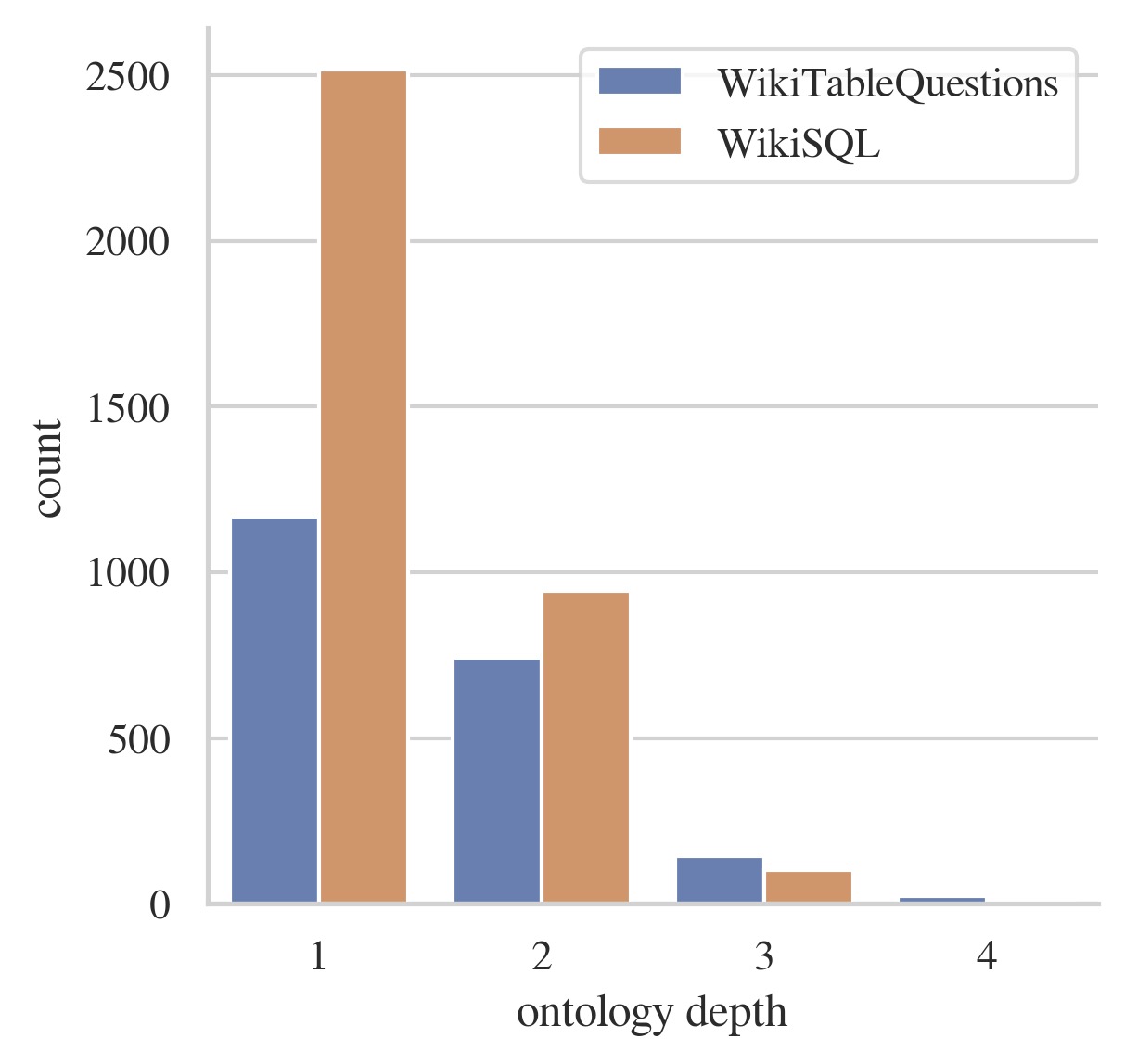}
  \caption{Distribution of column ontology depths in the WikiTableQuestions and WikiSQL samples in \ours~v1.1.1.}
  \label{fig:ontology_distribution}
\end{figure*}


\begin{figure*}[t!]
\scriptsize
\begin{verbatim}
    <entry category="MISC" eid="Id5" size="3">
      <modifiedtripleset>
        <mtriple>Apertura 2006 | JORNADA_OR_OTHER | Semifinals Ida</mtriple>
        <mtriple>Semifinals Ida | AWAY_TEAM | América</mtriple>
        <mtriple>Semifinals Ida | HOME_TEAM | Chivas</mtriple>
      </modifiedtripleset>
      <lex comment="WikiTableQuestions" lid="Id1">
          Chivas and América will compete in the semifinals of the Apertura 2006 tournament.
      </lex>
    </entry>

    <entry category="MISC" eid="Id76" size="6">
      <modifiedtripleset>
        <mtriple>Terry Jenkins | ROUND | 1st Round</mtriple>
        <mtriple>Terry Jenkins | YEAR | 2014</mtriple>
        <mtriple>[TABLECONTEXT] | [TITLE] | PDC World Darts Championship</mtriple>
        <mtriple>1st Round | OPPONENT | Per Laursen</mtriple>
        <mtriple>1st Round | RESULT | Lost</mtriple>
        <mtriple>[TABLECONTEXT] | PLAYER | Terry Jenkins</mtriple>
      </modifiedtripleset>
      <lex comment="WikiTableQuestions" lid="Id1">
          Terry Jenkins lost the game with Per Laursen in 
          the 1st Round of 2014 PDC World Darts Championship
      </lex>
    </entry>
\end{verbatim}
\caption{Examples of DART instance}
\label{fig:instance_examples}
\end{figure*}


\begin{figure*}[ht!]
 \centering
 \includegraphics[width=0.95\textwidth]{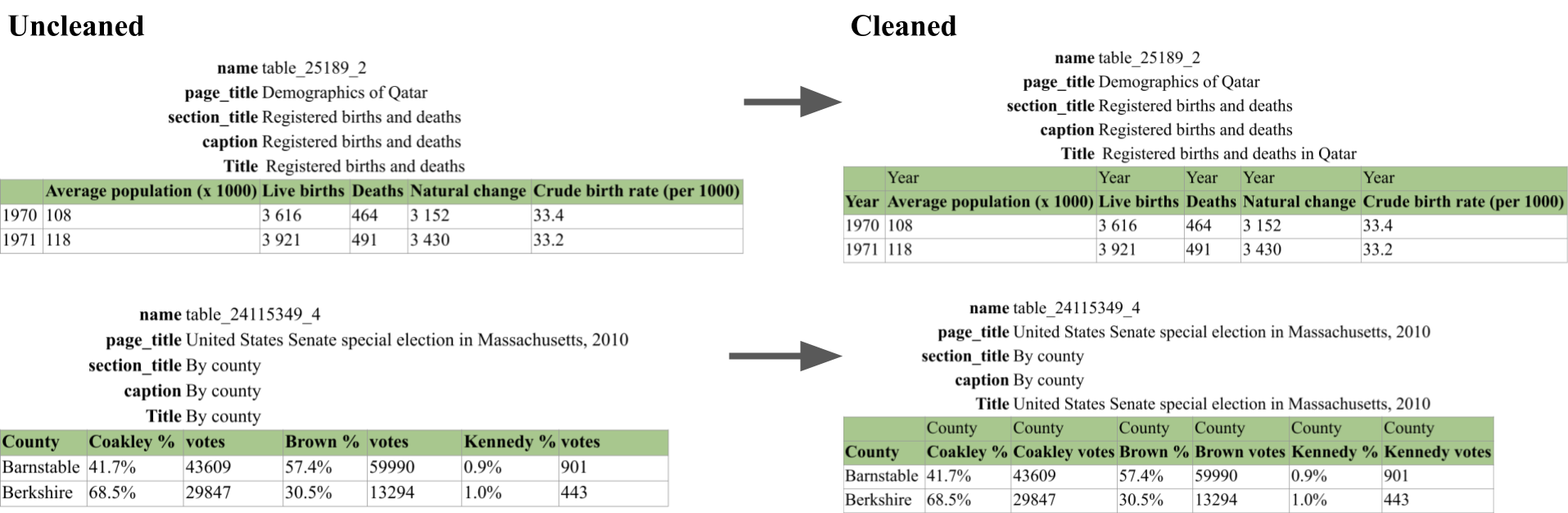}
 \caption{An example of the data cleaning. The top left table had a missing column name and the table title was not specific to the data; our internal annotators add the missing column name ``Year'' and linked the rest of the columns to the ``Year'' column. The bottom left table had repeat column names in the table; our internal annotators disambiguate the columns by making the column names more specific. }
 \label{fig:malformed_example}
\end{figure*}


\begin{figure*}[t!]
  \centering
  \includegraphics[width=0.85\textwidth]{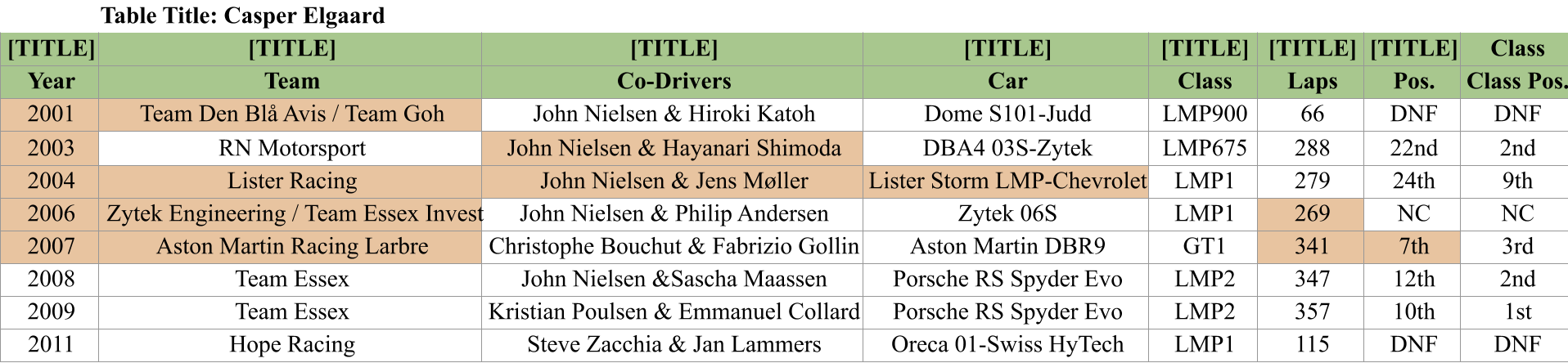}
  \caption{A WikiTableQuestions table that uses [TITLE] in the ontology.}
  \label{fig:title_example}
\end{figure*}


\begin{figure*}[t!]
  \centering
  \includegraphics[width=0.7\textwidth]{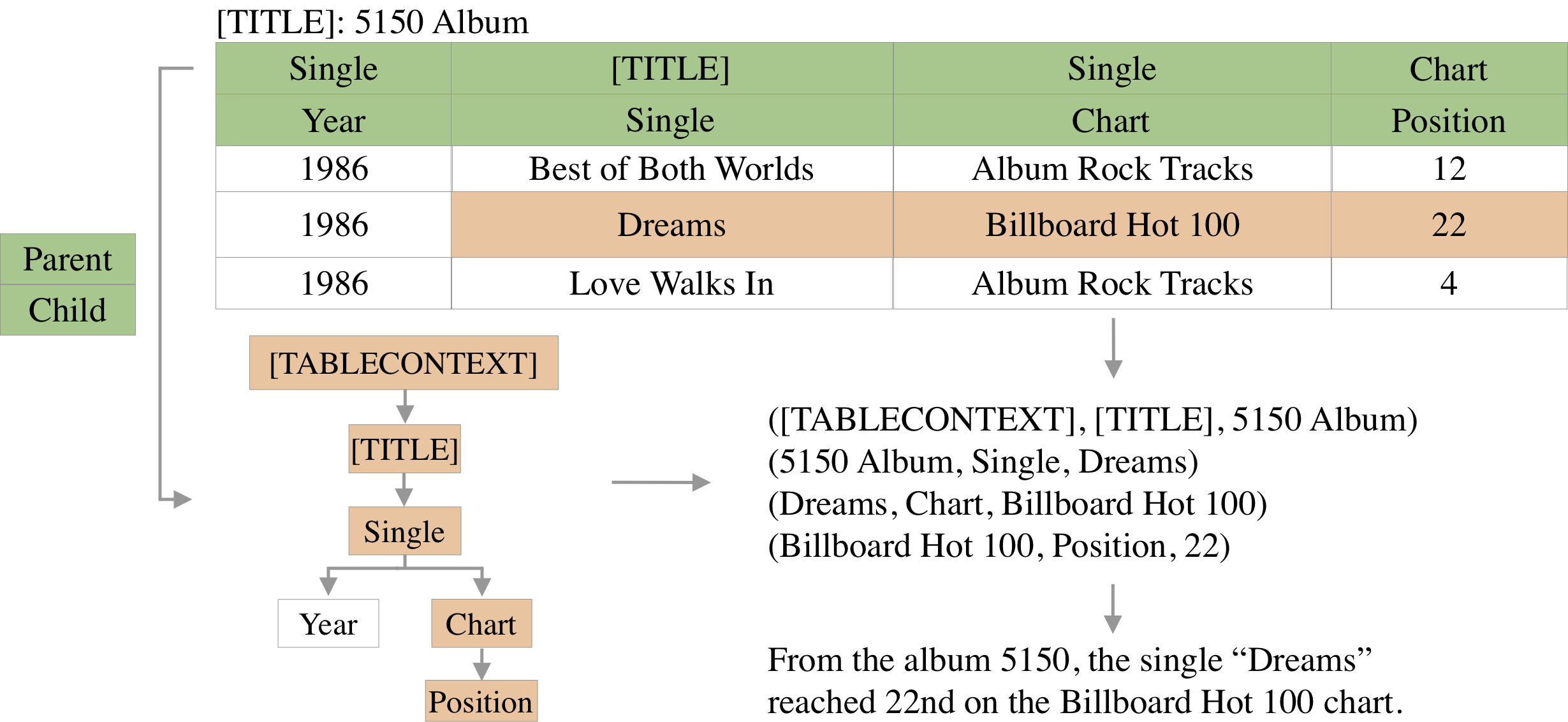}
  \caption{A manually annotated table from WikiTableQuestions with a sentence that uses the table title.}
  \label{fig:human_annotation_two}
\end{figure*}


\begin{figure*}[t!]
  \centering
  \includegraphics[width=1\textwidth]{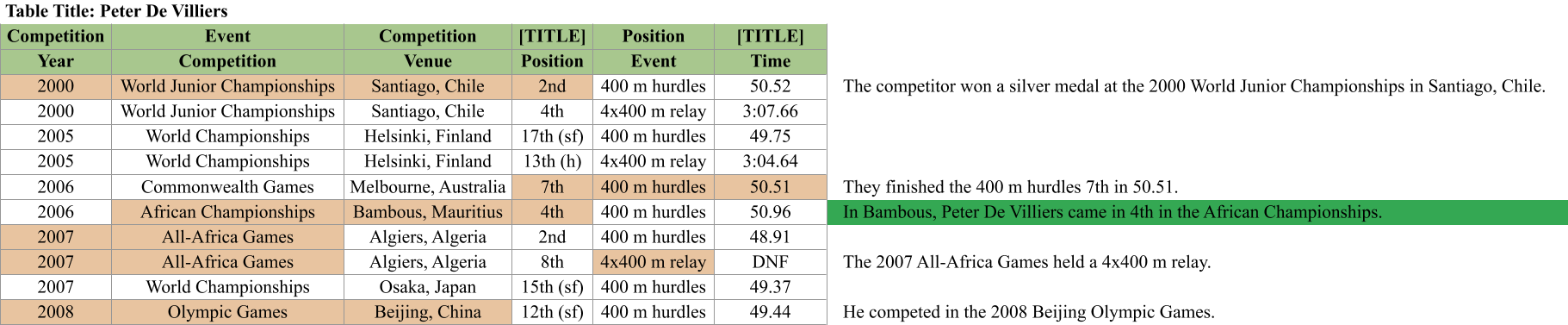}
  \caption{A manually annotated table from WikiTableQuestions. Annotators created a table ontology, and they wrote sentences
  encapsulating the information in the orange cells for a given row. Whenever a sentence referenced the table title, that sentence was also highlighted green.}
  \label{fig:WTQ_Example}
\end{figure*}


\begin{figure*}[t!]
  \centering
  \includegraphics[width=.7\textwidth]{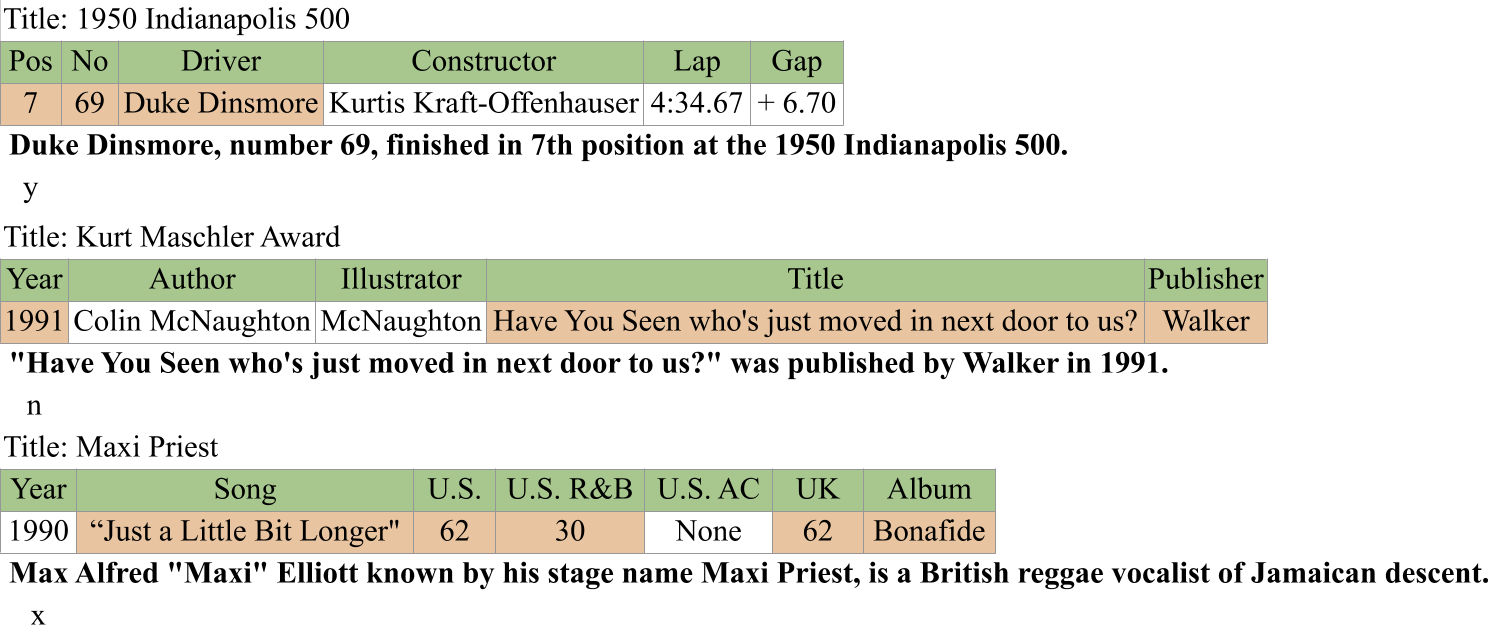}
  \caption{An example of collected MTurk-generated sentences for WikiTableQuestions. Internal annotators went through the generated sentences and checked for both sentence coherence and title usage. Below the generated sentences, `y' meant the sentence references the table title, `n' meant the sentence did not use the table title, `x' meant the sentence was nonsensical.}
  \label{fig:MTurk_example}
\end{figure*}


\begin{figure*}[t!]
  \centering
  \includegraphics[width=.7\textwidth]{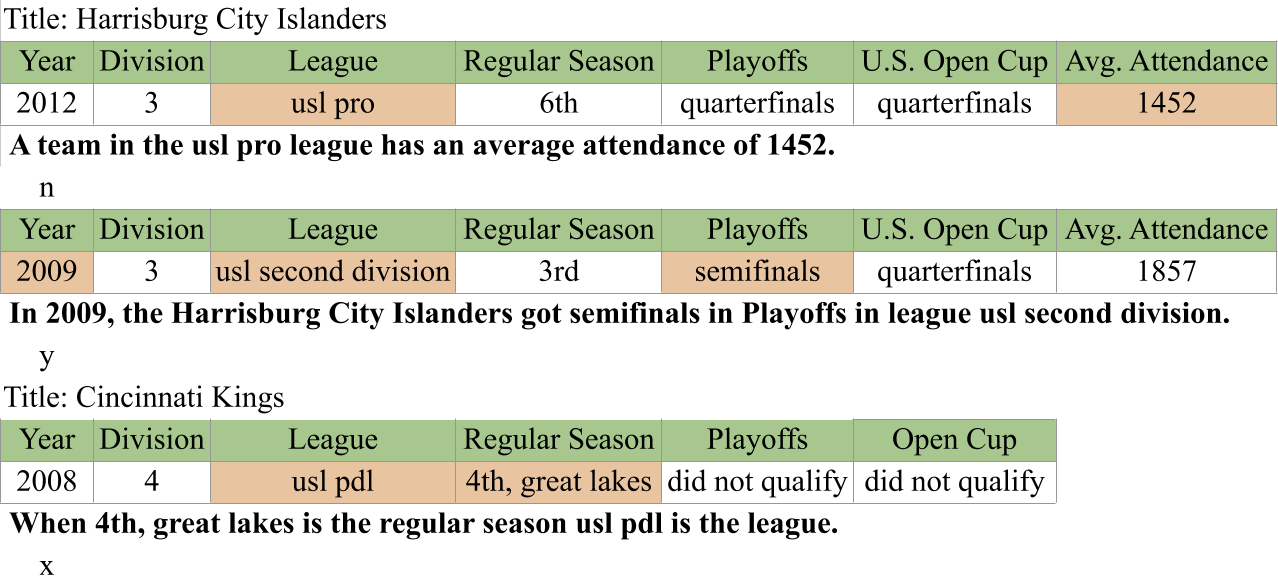}
  \caption{Automatically generated declarative sentences from WikiSQL with human validation. Annotators went through the generated sentences and checked for both sentence coherence and title use. Below the generated sentences, `y' meant the sentence references the table title, `n' meant the sentence did not use the table title, `x' meant the sentence was nonsensical.}
  \label{fig:WSQL_Declarative_Example}
\end{figure*}


\begin{figure*}[t!]
\scriptsize
\begin{verbatim}
    - Sample 1 -
    Input triples:
    <H> Peru Earthquake <R> scale of disaster <T> 250k homeless 
    <H> Peru Earthquake <R> year <T> 2007

    BART-base output: 250k people were killed in the 2007 philippine earthquake .

    - Sample 2 -
    Input triples:
    <H> [TABLECONTEXT] <R> game <T> 3 
    <H> 3 <R> attendance <T> 10 637 
    <H> [TABLECONTEXT] <R> [title] <T> 2006 Minnesota Swarm season
    
    BART-base output: the minnesota swarm played in front of a crowd of 10 , 684 people .

    - Sample 3 -
    Input triples:
    <H> Andrew Phelps McCormick <R> state <T> TX 
    <H> Andrew Phelps McCormick <R> active <T> 1892-1916
    
    T5-base output: andrew phelps mccormick was active from 1892 to 1616 in texas .



\end{verbatim}
\caption{Examples of hallucinated outputs of pretrained models trained on DART}
\label{fig:hallucination_examples}
\end{figure*}


\begin{figure*}[t!]
\scriptsize
\begin{verbatim}
    - Sample 1 - 
    Input triples:
        <H> Andrew Rayel <R> associated Band/associated Musical Artist <T> Christian Burns
        <H> Andrew Rayel <R> associated Band/associated Musical Artist <T> Jonathan Mendelsohn

    reference: 
    andrew rayel , is associated with musical artist jonathan mendelsohn and christian burns .
    
    train on WebNLG - BART-base output: 
    christian mendelsohn and andrew rayel are both associated with the same band , christian burns .
    
    train on DART - BART-base output: 
    andrew rayel is associated with christian burns and jonathan mendelsohn .
    
    - Sample 2 -
    Input triples:
    <H> Indie rock <R> stylistic Origin <T> New wave music

    reference: the stylistic origin of indie rock is new wave music .

    train on WebNLG - BART-base output: 
    the alternative rock genre is new wave .

    train on DART - BART-base output: 
    indie rock is influenced by new wave music .
    
    - Sample 3 -
    Input triples:
    <H> Abradab <R> associated Band/associated Musical Artist <T> Magik rapper
    <H> Abradab <R> associated Band/associated Musical Artist <T> Kaliber 44

    reference: 
    abradab , an artist for the band kaliber 44 , is associated with magik ( rapper ) .

    train on WebNLG - BART-base output: 
    magiber 44 is the creator of abradab , which is also associated with the magik rapper .

    train on DART - BART-base output: 
    magik rapper and kaliber 44 are the associated musicians of abradab .

    
    - Sample 4 -
    Input triples:
    <H> Alfa Romeo 164 <R> assembly <T> Milan
    <H> Alfa Romeo 164 <R> related Mean Of Transportation <T> Saab 9000

    reference: 
    the alfa romeo 164 , which is assembled in milan , is a related means of transportation to saab 9000 , 
    in that they are both cars .

    train on WebNLG - T5-base output: 
    alfa romeo 164 is a transport vehicle for saab 9000 and is found in milan .

    train on DART - T5-base output: 
    alfa romeo 164 ( assembled in milan ) is a related transport vehicle to saab 9000 .
    
    - Sample 5 -
    Input triples:
    <H> Akeem Ayers <R> former Team <T> Tennessee Titans
    <H> Akeem Ayers <R> draft Pick <T> 39

    reference: 
    akeem ayers ' former team was tennessee titans and he was number 39 in the draft pick .

    train on WebNLG - T5-large output: 
    akeem ayers was drafted with the 39th pick by the tennessee titans .

    train on DART - T5-large output: 
    akeem ayers , a former player of the tennessee titans , was the 39th draft pick .


\end{verbatim}
\caption{Examples of model outputs - with or without DART data augmentation}
\label{fig:data_augmentation_output_examples}
\end{figure*}

\end{document}